# Software Toolkit for Building Embedded and Distributed Knowledge-based Systems

Dmitri Soshnikov

Department of Computational Mathematics and Programming,

Moscow Aviation Institute (Technical University)

Moscow, Russia

dsh@freenet.ru

**Abstract**

The paper discusses the basic principles and the architecture of the software toolkit for constructing knowledge-based systems which can be used cooperatively over computer networks and also embedded into larger software systems in different ways. Presented architecture is based on frame knowledge representation and production rules, which also allows to interface high-level programming languages and relational databases by exposing corresponding classes or database tables as frames. Frames located on the remote computers can also be transparently accessed and used in inference, and the dynamic knowledge for specific frames can also be transferred over the network. The issues of implementation of such a system are addressed, which use Java programming language, CORBA and XML for external knowledge representation. Finally, some applications of the toolkit are considered, including e-business approach to knowledge sharing, intelligent web behaviours, etc.

## 1 Introduction

Nowadays, the largest sector of information technology is oriented towards networking environments. Since people mostly work in groups, you will not see a stand-alone computer in the enterprise nowadays: all software systems that support collective work processes naturally operate in computer networks.

Therefore, a special emphasis must be placed on moving the traditional information technologies and approaches towards networking model and open heterogeneous environments. While fundamental and widely used areas like database and information systems design receive enough attention, the other areas are less developed. One of such areas is the design of **knowledge-based systems** and **distributed expert systems**, which is one of the approaches to sharing and collectively using knowledge over computer networks.

Another aspect important for knowledge-based systems is their integration with traditional information systems. An expert system with it's own interface designed only for user interaction would not be useful in a real-life situation, where most of the information must come from the already existing database. Widely used approach to develop knowledge-based system that will be a part of a larger information system would be in developing the whole system from scratch in a high-level programming language, because traditional tools like CLIPS do not provide effective means for their integration into other software products.

Therefore it seems important to develop a software toolkit for creation of knowledge-based systems which are able to be integrated into larger information systems, and to interact with such different systems over computer networks. In the previous paper presented on CSIT'99 such an approach was discussed with respect to production expert system shell with somehow simplified rule syntax and static knowledge representation in form of attribute-value pairs. In this paper we apply the same approach in developing a toolkit for creating more complex system based on frame knowledge representation with production rules.

## 2 Distributed and open knowledge-based systems

### 2.1 Technologies for buiding distributed and open software systems

In the field of modern information technology there is large tendency towards more open software systems and enterprise models for building large-scale software products. Many underlying technologies for creating distributed systems has been developed, including such standards as CORBA [8], Enterprise JavaBeans, RMI [10], DCOM, etc., however, they are still to be widely used in practical applications.

Most of the technologies follow the **component approach** whereas the complex system can be built from smaller functional blocks (components, CORBA or COM/ActiveX objects, JavaBeans), which can interoperate either on one computer or seamlessly over the network. Thus the systems become more open, because different components can be combined provided they follow the same **interfaces**, and it is also possible to construct **distributed systems**, with components running on different network nodes. Moreover, since the components can be developed in different programming languages and used on different platforms, it is possible to construct complex heterogeneous environments.

In a way, **Java technology** is the most interesting and advanced among the emerging technologies for the enterprise. It supports **CORBA** — one of the major standards in this area — as well as it's own JavaBeans component model and related RMI technology. Java programming language and related technology is highly standardized, and allows creating reusable and cross-platform libraries and applications with less programming effort. Dynamic nature of the language and particularly Java Reflection API allow to create flexible and open applications even on a smaller scale without using networking component models.

### 2.2 Approaches and tools for knowledge-based systems

Let us briefly cover some of the software systems that apply principles mentioned above to the knowledge-based systems.

One of the earliest attempts is the system called **BOW** [3], which is basically the expert system shell designed as software library which can be called from any other program to load the specified textual knowledge-base in the form of production rules and perform the reasoning using backward chaining. A special module was also designed which allowed the system to be used in the networking environment as a reasoning server, with all calls to it seamlessly redirected by the RPC protocol. In this way the knowledgebase was located on the server where the reasoning process also took place, and only static knowledge about the problem being solved was transferred over the network.

The principle of distributing knowledge in production knowledge-based system was further developed in **DIET toolkit** [1], in which the component approach was applied to separate different components of an expert system (inference engine, knowledgebase, static and dynamic knowledge sources, etc.). Thus with this system it was possible to build configurations where either static or dynamic knowledge (or both) are transferred over the network in the internal representation. CORBA was used as the basic component interaction protocol, which made it possible to use the system as a reasoning server and make calls to it from other software information systems.

Another approach for embedding reasoning into larger software systems used in the DIET toolkit was **high-level code generation from the knowledgebase** [2], in which the programming code in a high-level programming language (Pascal, C, Java) is automatically generated from the knowledgebase. The generated code consists of a set of recursive procedures (one for each attribute used in the static knowledge representation), and, when executed, models the backwards chaining reasoning with simple "first-come" conflict resolution strategy.

A good example of embeddable technology is **AMZI Prolog** [5], which is a fully-functional Prolog system available as ActiveX component. It can therefore be called from any other software system which supports ActiveX to perform knowledge-based tasks, which is a good solution for embedding logic components into information systems. However, using Prolog instead of some other knowledge-representation scheme limits the possibility of creating distributed systems which dynamically exchange knowledge between nodes, and also reduces the feasibility of developing knowledge-based systems by not computer experts.

Finally, it is worth to mention a project called **JESS** [6], an expert system shell written in Java, which supports subset of CLIPS, a very widely known expert system shell. Being distributed as a set of Java classes, it can in principle be used as a part of larger system, although it does not provide clear API to do so. It is however ready to be used for remote consultation through the Internet in the form of a Java applet.

### 2.3 How it is related to Distributed Artificial Intelligence

Nowadays the most popular approach to **distributed artificial intelligence** (DAI) is the concept of **intelligent agents**, particularly **mobile agents** — the complete atomic intelligent systems which are set to accomplish a specific task by means of their internal intelligence. Such an agent is a knowledge-based system in itself, but is typically capable of sharing external static knowledge with other systems over the network, and work cooperatively with other agents towards a larger common goal.

While mobile agents typically transfer their code to other network nodes to be executed there, the component approaches presented above all imply that components run on the fixed node and only exchanges information with other components. But components are able to exchange both static and dynamic knowledge (i.e. share their reasoning logic with each other), which agents normally do not (in agents the notion of environment is clearly separated from internal reasoning mechanisms).

Agent and component approaches normally target slightly different problems, and therefore we cannot really talk about which one is beneficial. In the author's opinion the component approach addresses the real-life problem of using knowledge-based systems in networking environments in a straight-forward manner (which is a further development of traditional client-server paradigm), while agent theory is better developed in theoretical terms and is targeted towards future developments in theoretical and practical artificial intelligence.

## 3 Architecture of the Toolkit

In [1], the architecture of the toolkit for creating distributed knowledge-based systems was outlined. It was based on somehow simplified knowledge representation in the form of production rules and attribute-value pairs. Our further goal was to extend this approach for more complex **frame-based knowledge representation scheme**, which would also allow for more transparent integration of the intelligent component with other models (mainly procedural and object-oriented languages and relational databases).

### 3.1 Basic principles

The proposed system architecture is shown on Fig. 1. It is structured as a set of Java classes, corresponding API, and CORBA objects for accessing the system remotely.

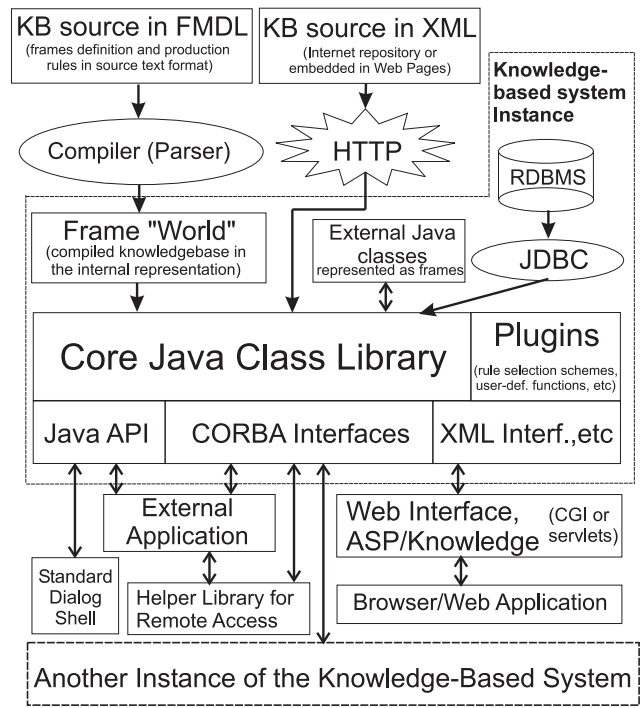

Figure 1: Architecture of the Toolkit

The heart of the system is the **core Java class library**, which is responsible for the inference process and the overall functioning of the system. Everything above the core class library on Fig.1 shows how the system obtains knowledge and data necessary for its operation, everything below — how the library can be interfaced and called from inside other applications or interactively. The figure also shows that one instance of the system can interface with another one via CORBA interfaces to exchange knowledge.

### 3.2 Knowledge representation and inference

The system uses frame knowledge representation model and production rules for formulating dynamic knowledge. The basic unit of knowledge is a **frame**, which consists of a set of **slots** that can hold **values**. The basic **data types** used are **scalar**, **list**, and **reference**; scalar data type is further subdivided into **integer**, **boolean**, and **string**. Richer set of data types (including **floating-point** values, **text** objects representing whole paragraphs of text of web pages, etc.) can be considered for further development.

Each slot has a fixed associated data type, and may also have **default value**, a set of **constraints** and **actions**, which can be fired when a value is assigned to the slot, or when the value of the slot is required.

Frames form the **inheritance hierarchy**, where each frame can have no more then one parent. Multiple inheritance presently cannot be used, but is considered for further development. Inheritance is **dy-** **namic**, i.e. the parent of a frame can be determined at run-time by either assignment to the `parent` slot as a result of a rule execution, or by **frame specification**, where the parent is determined based on slot constraints in a certain subtree of inheritance hierarchy.

The knowledge for the actual knowledge-based system is formulated in a high-level descriptive language (which we call **Frame Model Description Language**, or FMDL), which combines both static knowledge on problem domain (the frame hierarchy and slot properties) and the dynamic knowledge in the form of production rules. This language description is then translated into a collection of Java objects representing frames (so-called **Frame World**), which can be serialized and later easily loaded for the actual use.

Another way to store and use knowledge model is by using specific XML [11] representation, which is also directly understood by the core library. By using this XML representation the model or a certain part of the knowledgebase can be obtained from the remote sources via HTTP protocol (eg. from the Internet knowledge repositories, from knowledgebases embedded in web pages, or from other intelligent systems which support XML) and integrated into the frame world with which the system operates. Using XML from knowledge interchange has many advantages, one of them is being able to transform different XML representations from different systems via XSL and thus allowing two systems with slightly different XML knowledge representations to interact.

All frames in the "Frame World" are represented by specific Java classes, descending from the abstract `Frame` class. In this way it is possible to operate on frames of different kinds, for example, located on different network nodes (see section 3.3 below) or representing Java class (section 3.4).

All production rules that come from either sources are translated into slot actions which fire when the value of the slot is needed for inference (rules for backwards inference), or when the slot value is changed (rules for forward inference). All expressions that are part of rules (arithmetic, logical, etc.) are stored as their tree representations. Some other actions (besides rules) can be assigned to slots as well, specified, for example, as external Java functions or CORBA calls. This is one simple way to extend the pure knowledge-based inference model with procedural functions. Another commonly used action would be questioning the user about the value of the specific slot.

When the value of a slot is to be determined, all actions specific to this slot are considered in the **conflict resolution process**, which is determined by the specific Java plugin class. The default behaviour is to try the first available action first, but other behaviours (eg. most complex rule is applied first) can be assigned by setting a specific plugin for conflict resolution in the runtime from the standard set of plugins, or by writing a custom plugin in Java. It is also possible to assign conflict resolution plugins separately for different frames. In this way it is possible to customize the particular type of backwards inference applied to the frame-production model.

When no rules for a specific frame can be applied, all actions/rules for parent frame are checked in the same manner. When finally a certain rule applies (i.e. it's left-hand-side returns a true value) the value obtained is assigned to the slot of the original frame (in this manner the polymorphism is implemented), and "on change" actions are performed, in the precedence determined by another customizable conflict resolution algorithm, thus applying one or all the applicable forward-chaining rules.

Forward chaining and other inference techniques are not yet implemented in the experimental prototype. One of such techniques is **frame specification** mentioned above, when the frame parent is determined based on the slot constraints; another is **implicit search in place of existential quantification**, in which all children from the subtree of a certain frame are tested to obey certain condition in the rule premises until the suitable one is found. Such techniques, although slowing down the inference significantly, would greatly enrich knowledge representation used and enhance it further with respect to single-object systems with attribute-value pairs.

### 3.3 Distributed frame hierarchy and remote rules

One of the main features of the toolkit is the ability to be distributed over the network and share different knowledge types. This is implemented through two principles: **distributed frame hierarchies** and **remote rules**.

We will call a set of Java classes including core library and "Frame World" representation running on a particular network node **an instance**. Distributed knowledge processing and sharing takes place when different instances exchange static or dynamic knowledge between each other, thus solving one problem with one frame hierarchy.

Fig. 2 shows the example of distributed frame hierarchy, where different frames are located on different machines, and only exchange static knowledge (frame attribute values) between each other. Root frame called `Object` resides on one machine, and others refer to it and to other remote frames by using special Java classes — **frame stubs** — in their "Frame World" representations. When the local inference process requires

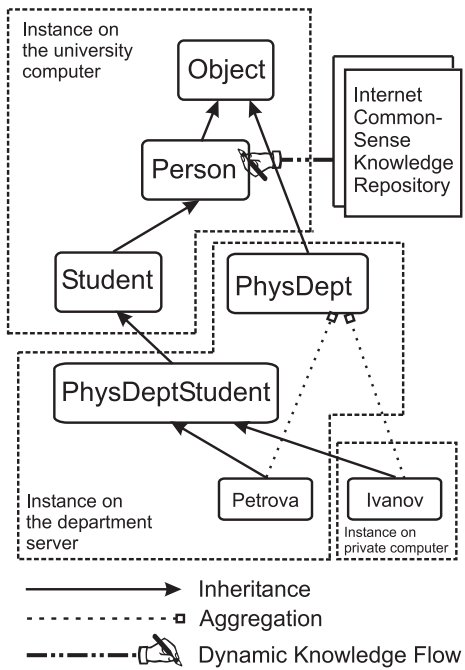

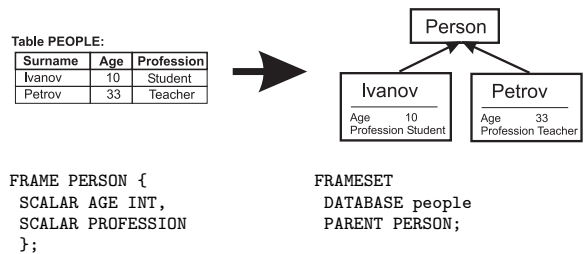

```
FRAME PERSON {                 FRAMESET
  SCALAR AGE INT,                DATABASE people
  SCALAR PROFESSION              PARENT PERSON;
};
```

Figure 3: An example of database-generated frame hierarchy

tributed manner on different network nodes.

Figure 2: An example of distributed frame hierarchy
a particular slot value from a stub, it forwards the call through CORBA interface to the remote instance, causing the inference process to start there until the required value is obtained. CORBA object reference of the calling frame is passed to the remote parent, so that all slot values are taken from the original frame and polymorphism is ensured. Thus the remote frame is transparently used in the local inference process, and knowledge engineer designing the knowledgebase for a particular instance does not need to know the internal logic of the remote frame, just slot names.

It can be noted, however, that in the distributed frame hierarchy the inference process takes place on different computers. In principle this architecture can utilize **parallel inference**, where all the values needed at a certain point are inferred simultaneously in concurrent threads or on different servers. This however has not been implemented yet, because, as with any concurrent processes there are many complications that arise; however, the extension to the parallel inference should not be too complex in the presented architecture.

To speed up the process and minimize the network bandwidth for inter-instance communication each frame stub also keeps a cash of obtained values, so that there is no need to re-send them through the network when they are needed more then once (which is often the case in a complex knowledgebase).

In distributed frame hierarchy the frame inheritance tree is the same for the whole group of instances, which we will call **a cluster**, only static knowledge is exchanged, and inference process takes place in the dis-

However, in some situations it is desirable to be able just to take dynamic knowledge (production rules) from the remote network location and apply them on locally running instance. This is for example true for **knowledge repositories**, when we want just to place a certain knowledge for others to use, not providing any inference services. In this case a frame attribute on one machine requires a remote rule for it's inference, in which case the rule representation would be transferred over the network (the dynamic knowledge) and used in inference process by the local inference engine.

Of course, the system can use a combined distribution topologies, where different complex knowledge exchange strategies co-exist in one configuration. It is important to point out that frame-based approach allows to control the distribution strategy in much more flexible way comparing to previously suggested single-object approach [1].

### 3.4 Integration with procedural languages and relational databases

In creating real-life expert systems, it is important to be able to use external databases as source of data for solving the problem. Frame-based design allows for easy integration of knowledge-based system with relational databases in different ways, among which

- Representing database table as a set of frames with a common parent. An example of this situation is shown on Fig.3. A set of frames generated from the table is called **frameset**, and it is represented in the "Frame World" as one Java class. Rows from the table are dynamically accessed when there is a need for a certain attribute from one of the frames in the frameset. Rules cannot be further assigned to the frameset-generated frames, but can be to the parent frames.

- A **system frame** for accessing one specific row from the table based on SQL query. **Frame generator** statement retrieves a specific row from the table and creates the corresponding frame in the "Frame World". Such generated frame can have further-assigned rules and actions.

- A **system action** for retrieving the value from the database based on the SQL query. This is probably the simplest form of database access, which has also been used in non-frame systems, but it can only be used to retrieve one value and not the frame structure (although this value can be a list).

Frame model also allows to interface object-oriented programming languages (particularly Java) in a consistent manner, by having an arbitrary Java class exposed as a frame. For each such class there is a special **stub frame** (similar to remote stub frame), which, when it's slot is being accessed, calls the method with the same name of the underlying Java class. Java Reflection API allows the stub to see which class fields and methods are available. The naming conventions are those used in JavaBeans; in fact, Java class being exposed as a frame can also be a JavaBean.

There is a number of specific **system frames** to be developed in Java which can serve certain procedural tasks or be used as intermediate gateways. One example is query-based database access mentioned above, or the mechanism for **web data gathering**, which allows to collect data from web pages based on certain templates dynamically in the process of inference.

Apart from interfacing arbitrary Java classes as frames, many other components of the toolkit can be extended by writing Java code. One is the above-mentioned possibility to write arbitrary conflict resolution algorithms, another is a special API for extending the built-in functions with arbitrary Java functions. All those things allow to integrate the core Java class library with custom applications very tightly, creating mixed procedural and knowledge-driven applications for solving complex tasks.

### 3.5 Interfacing the toolkit

There are several ways the toolkit can be called from other applications or used interactively:

- Using **Standard Dialog Shell** (see Fig. 1) the toolkit can function as a stand-alone expert system shell with it's own interface.

- Calling core Java class library directly from within any Java application via special API. Note that the knowledgebase used can be written beforehand in FMDL and translated into internal serialized representation, so that the application in the simplest case needs just to load it from any stream (from disk file, Internet connection, etc.) and start inference. Knowledge can also be loaded dynamically in XML representation.

- Application written in any other programming languages that support CORBA can call the toolkit instance via exposed CORBA objects. In this case there needs to be a knowledge server (or a cluster) running the instance of the knowledge-base (or distributed frame hierarchy) which the application will access for performing it's intelligent activity.

- Languages that do not have CORBA natively built in (eg. older versions of Delphi, C, etc.) in Microsoft Windows can use the special **DLL helper library** provided for communication with the toolkit. Also specific components (Delphi components, ActiveX, JavaBeans, etc.) can be built to access the system from other standard environments.

- There is also a **web interface** planned for development for accessing the system through the web, in the form of Java Servlet. Servlet technology is very suitable for implementing intelligent systems, as it allows the state of the web application to be completely preserved over invocations. On the other hand, traditional web technology can also be used, as the system state can be serialized at any time and restored at next invocation from the client.

### 3.6 Summary

As it was briefly mentioned above, the proposed architecture has been designed with the following major goals in mind:

- To allow flexible interaction with other software systems on different levels, including the ability to call the inteligent system from an arbitrary software system to perform inference, and vice versa, to call the procedural modules from within the inference process of the intelligent system.

- To allow distribution of knowledge among different network nodes and exchange of both static and dynamic knowledge depending on the nature of a problem being solved.

Of course, there are other goals that apply to any intelligent system, like rich yet clear knowledge representation, effectiveness, etc.

Now that the architecture of the toolkit has been described, it becomes clear that frame knowledge representation is very suitable for achieving the above-mentioned goals. Namely, the following advantages of using frame knowledge representation can be noted:

- Frame knowedge representation is very similar to traditional object-oriented approach, which allows to use frame-like structure as a "common denominator" when interfacing the intelligent system and any other software systems. Namely objects from any open software system (Java Classes, JavaBeans, CORBA or COM Objects, etc.) can be exposed as frames to the intelligent module, and vice versa any frame withing the knowledge model can be handled as an externally-accessible object by the calling software system.

- Frames can also be effectively used to provide interfaces with relational databases (as it was outlined in section 3.4) and other types of structured information (Web information gathering, etc.).

- Frame knowledge model provides a natural way of clustering knowledge and particularly dynamic rules around frames (as it was mentioned in section 3.2, all rules belonging to one frame are grouped together in the internal representation), which in turn provides the natural way of distributing knowledge among network hosts. Distributed frame hierarchy and remote rules allow to construct quite complex distribution strategies for both static and dynamic knowledge sharing, which nonetheless in many cases naturally correspond to the knowledge model being created. The issue of developing distributed knowledgebases has to be considered separately, but it is quite clear that frame-based models can be based on already-developed methodologies for object-oriented systems.

## 4 Practical Applications of the Toolkit

The presented toolkit architecture provides a way for different applications to use the process of distributed inference, thus embedding complex knowledge-based functionality. Although the number of potential applications can be very big, below we discuss some of the possibilities.

### 4.1 Knowledge-based Components for Information Systems

As it was pointed out above, the toolkit allows for very tight integration of information and intellectual components in the software system, where both procedural modules can call the toolkit core library to start logical inference, and frame model can access procedural modules when required by the inference process, not to mention access to other data models like relational databases and web data gathering.

An intelligent information system based on the described toolkit (although in slightly less general form) has been created and is now used for diagnosing patients with benign prostatic hyperplausea at the Botkin State Hospital, Moscow [4]. The system is created in Borland Delphi, and utilises both high-level code generation approach for built-in knowledgebase module, and the ability to call the external knowledgebase server running the Java version of the toolkit with single or distributed knowledgebases. Although the ability to create distributed knowledge repository is not yet fully exploited due to the limited hardware of the hospital, we feel that the approach itself can be applied to this situation very successfully.

For interfacing the knowledge server the system uses Delphi component developed around the DLL, which makes CORBA calls to the server. The DLL itself was developed in C++. Although latest versions of Delphi support CORBA natively, this approach allows to use the access to knowledge server into older versions as well, and also provides an extra abstraction layer for switching to another protocol, eg. XML-RPC [11].

### 4.2 Collective Knowledge Sharing

One of the immediate usages of distributed knowledgebases is collective knowledge sharing, where different knowledgebases can reside on different computers on the network and be used collectively to obtain the solution to a certain problem. Since the inference can also take place on different hosts, concurrent inference can be used to speed up the process by splitting the inference through different network hosts.

Creating distributed knowledgebases can be very convenient in many situations, because each knowledgebase component can be maintained and kept on a separate computer, and used in consultations. Since it is now recognized that knowledge is a good which can be sold, this approach allows also to actually charge external users on "per-consulation" basis without giving out the actual knowledge. In addition, knowledge repositories allow charging on the basis of "per-rule" used in inference. In the growing popularity of **e-business** and **Business-to-Business** solutions such an approach for selling knowledge becomes very important. The system can be easily extended to track all the requests and use some authentication mechanism.

Such distributed knowledgebases can be successfully used in **virtual corporations**, which produce some intellectual products largely based on knowledge. For example, the corporation which suggests the best place in the world for a person to go for a particular reason (on vacation, on working contract or to relocate permanently) can use separate knowledgebase for different countries which are maintained separately and provide specific to that country advice based on the common premises.

Developing distributed knowledgebases is a separate issue in itself. While there are no specific methodologies for knowledgebase development in general, separating a knowledgebase into different parts requires even more effort and consideration. Some special cases when it is relatively easy and practical to create distributed knowledgebases are considered in [1], one example being the case with **separable knowledge subdomains**, where there are several knowledgebases for slightly different knowledge domains (and therefore with mostly different sets of attributes) that intersect only on small number or attributes, which can be maintained separately. In fact, frame representation naturally separates dynamic knowledge with respect to target frames, thus creating some subdomains which can be used as a basis for separation.

However, frame model creates more cases where knowledge distribution can be exploited, because of more complex nature of frame interrelations. However, it also requires some better-developed techniques for collective development of such knowledgebases, and even some software systems to support them. Since all parts of distributed knowledgebase have to work cooperatively, it is important to ensure that at least common attributes are named exactly the same and form the common ontology. Creating a tool for cooperative development of ontology and taxonomy of frame model, with direct interface to the presented toolkit is an interesting issue, as well as the development of appropriate methodologies. Those issues would probably need to be addressed separately and covered in a separate publication.

### 4.3 Remote Consultations

Remote consultation is the simplest form of distributing knowledge, where a single knowledgebase is used to solve a particular problem. The issue of remote consultations is addressed in [1] in more detail, and it should be noted that presented toolkit naturally supports remote consultations both for humans and software systems with different types of knowledge transfer. The toolkit includes user interfaces that can be used for accessing remote knowledge servers in the form of separate application, or Java Applets and ActiveX controls for consultation through the Web. There is also a Servlet system planned for development, which will perform all inference-related operations on the server, and only present the user with forms for answering questions generated by the system.

### 4.4 Intelligent Web Behaviours

Remote consultation is just one explicit way of using knowledge through the Internet. Knowledge-based technology can also be implicitly applied to other aspects of the WWW usage, some of which we will briefly mention below.

**Intelligent Navigation**

One of the problems of the current state of development of the WWW is the difficulty of finding needed information on the Web. Most of search engine technologies implement keyword search, which basically finds the pages which contain certain words. Some more complicated search engines use AI techniques for finding related pages, but at any rate everything is based on plain-text representation of the initial information.

One of the ways to deal with this situation is to somehow enrich the web contents with *knowledge* about the information presented in the page, and then use this knowledge to find the wanted pages based on reasoning. The process of implementing such a system on a world-wide scale seems quite unrealistic at the moment, but it may work for smaller intranets. The main principle is to have some common ontology for specifying the general direction of search, some common-sense knowledge server which will suggest a set of servers for each direction, and then to use the process of distributed inference based on the knowledge embedded in the web pages. Frame representation is ideal for such a system, because each page can be represented naturally as frame descending from another more general frame specific for web server.

**Intelligent Content Generation**

Another issue related to embedding knowledge in web pages is **intelligent content generation**, which is basically producing web resources that show different information on the page based on logical reasoning. Server-side remote consultation is actually a particular case of intelligent content generation, because it asks questions which it needs to proceed with inference, i.e. the actual web content that is returned to the user depends on reasoning process.

For embedding such knowledge into Web pages it makes sense to develop a technique similar to Active Server Pages (ASP) technology of embedding procedural code into HTML. However, while procedural ASP implementations are based on the fact that linear program code can be easily generated from ASP source and then executed to produce required output, the situation with intelligent ASP is more complex, because of the non-linear nature of inference process. Although the very basic syntax of ASP can be preserved (using `<%` and `%>` as HTML/target language switches), some slight modification are required to the syntax, and also session management becomes more complex, since a lot of information (the whole "frame world") needs to be stored for each particular HTTP session. Thus it is more practical to implement ASP/Knowledge technology not as a pre-processor, but rather as a servlet, which keeps running even between HTTP requests and has all the related information loaded.

## 5 Conclusions

The presented approach creates a basis for complex enterprise knowledge-sharing, and integration of knowledge-based component into traditional information systems. Moreover, it is important to create a toolkit based on the set of new technologies (CORBA, Enterprise JavaBeans, XML, etc.), which would make it usable in contemporary projects and interoperable with other systems based on the same open standards. We believe that it would allow knowledge-based systems to be used more effectively in the wider range of real-life situations, so that the benefits of this class of software systems are fully exploited.

Based on this approach the experimental prototype of the toolkit was created and used in developing real-life intelligent information system. This prototype needs to be further developed into the production system and to fully exploit the ideas presented in this paper. Some of the ideas mentioned require more theoretical development and investigation before the actual implementation, however, the present version of the toolkit can be used in building network-aware intelligent systems ranging from simple client-server expert systems to more complex systems with distributed inference and collective usage of knowledge. In particular, the presented technology allows considering machine representation of knowledge to be a good, which can be sold on the Internet through e-business mechanisms using different payment plans.

Having a tool for easier creation of intelligent components for complex information environments would encourage developers to adopt more intelligent behaviours for their information systems, thus producing higher quality and more usable systems for information management in the enterprise. While the agent approach is still being developed and is not yet widely used, we hope that this somehow straightforward and alternative architecture would find it's use for developing knowledge-based systems in the networking environments.

## Acknowledgements

I would like to express my gratitude to the department of Computational Mathematics and Programming of Moscow Aviation Technical University, and in particular V.E.Zaytsev for his firm scientific advice and leadership.

I would also like to thank the staff of the Botkin State Hospital, and particularly I. V. Lukianov for providing an interesting task to be solved using expert systems technology with the developed toolkit, and for all the support and dedication, which finally led to creation of useful information system with intelligent components that is now used in the hospital.

Special thanks go to my wife Larisa, who patiently coped with my busy nights in front of the computer screen.

## References

[1] Soshnikov D. An Approach for Creating Distributed Intelligent Systems. In J.-C. Freytag and V. Wolfengagen, editors, *Proceedings of the 1st International Workshop on Computer Science and Information Technologies*, Moscow, Mephi Publishing, 1998. pp. 129–134.

[2] Soshnikov D. Software Tools for Creating Embedded Production Expert Systems.In *Collected Abstracts of X Anniversary International Conference on Computational Mechanics and Modern Applied Software Systems*, Moscow, MSIU, 1999. – pp.325–326. (in Russian)

[3] Zaytsev V.E., Lukashevich S.Yu. Instrumental Tools for Building Embedded Expert Systems. Informatika, No. 3–4, 1991. pp. 30–40. (in Russian)

[4] Lukianov I.V., Zavedeev I.A., Soshnikov D.V.Using Expert Ananysis in Selecting Medical Treatment Tactics for Patients with Infravesical Obstruction. In O.B.Karyakin, K.M.Figurin, et.al., editors, *Collected Abstracts of Third Russian Scientific Conference with Participation of CIS Countries "Actual Methods for Treating Oncology Diseases in Urology.*, Moscow, Russian Oncology Scientific Center named after N.N. Blokhin, 1999. — pp. 131–132.

[5] Web Site of AMZI Prolog, `http://www.amzi.com`.

[6] Web Site of JESS, `http://herzberg.ca.sandia.gov/jess/main.html`.

[7] Russel S., Norvig P. Artificial Intelligence: A Modern Approach. Prentice-Hall, 1994.

[8] Orfali R., Harkey D., Edwards J. Instant CORBA. Wiley Computer Publishing, 1997.

[9] Minsky M. A Framework for Representing Knowledge. MIT, Cambridge, 1974.

[10] Java Technology Web Site, `http://www.javasoft.com`

[11] XML Web Site, `http://www.xml.com`